\documentclass[preprint,12pt]{elsarticle}

\usepackage{amssymb}
\usepackage{amsthm}

\usepackage[T1]{fontenc}
\usepackage{graphicx}
\usepackage{soul}
\usepackage{bm, makecell}
\usepackage{amsmath,amsfonts}
\usepackage{graphicx,pgfplots,tikzscale,pgfgantt}
\usepackage{arydshln}
\usepackage{bbm}
\usepackage{booktabs}

\usepackage[utf8]{inputenc}
\DeclareUnicodeCharacter{2212}{−}
\usepgfplotslibrary{groupplots,dateplot}
\usetikzlibrary{patterns,shapes.arrows}
\pgfplotsset{compat=newest}
\usepackage{subcaption}

\usepackage{algorithm}
\usepackage{algorithmic}

\graphicspath{{./fig/}}

\DeclareMathOperator*{\argmin}{argmin}
\newtheorem{definition}{Definition}

\begin{document}

\begin{frontmatter}

\title{Deep Feature Selection Using a Novel Complementary Feature Mask}

\author[label1]{Yiwen Liao}
\author[label2]{Jochen Rivoir}
\author[label2]{Rapha\"el Latty}
\author[label1]{Bin Yang}

\affiliation[label1]{organization={Institute of Signal Processing and System Theory, University of Stuttgart},
            addressline={Pfaffenwaldring 47}, 
            city={Stuttgart},
            postcode={70569}, 
            state={Baden-Wuerttemberg},
            country={Germany}}

\affiliation[label2]{organization={Applied Research and Venture Team, Advantest Europe GmbH},
	addressline={Herrenberger Strasse 130}, 
	city={Boeblingen},
	postcode={71034}, 
	state={Baden-Wuerttemberg},
	country={Germany}}

\begin{abstract}
Feature selection has drawn much attention over the last decades in machine learning because it can reduce data dimensionality while maintaining the original physical meaning of features, which enables better interpretability than feature extraction. However, most existing feature selection approaches, especially deep-learning-based, often focus on the features with great importance scores only but neglect those with less importance scores during training as well as the order of important candidate features. This can be risky since some important and relevant features might be unfortunately ignored during training, leading to suboptimal solutions or misleading selections. In our work, we deal with feature selection by exploiting the features with less importance scores and propose a feature selection framework based on a novel complementary feature mask. Our method is generic and can be easily integrated into existing deep-learning-based feature selection approaches to improve their performance as well. Experiments have been conducted on benchmarking datasets and shown that the proposed method can select more representative and informative features than the state of the art.

\end{abstract}


\end{frontmatter}


\section{Introduction}
\label{sec:intro}
Feature Selection (FS) is one of the most crucial techniques in machine learning and data science~\cite{guyon2003introduction,chandrashekar2014survey,li2017feature,cai2018feature,dokeroglu2022comprehensive}, attempting to select the most representative and informative feature subsets to reduce data dimensionality. Accordingly, the downstream learning algorithms can be more efficiently trained on the data with reduced features, while still maintaining reliable performance. Meanwhile, the computational and storage consumption are also reduced due to significantly less features. In addition, from the aspect of researchers and practitioners, feature selection has been frequently leveraged as a useful tool to identify the pivotal and representative features with the intention that experts can gain a better and deeper understanding of data, which plays an important role in practical and industrial scenarios~\cite{ding2005minimum,liu2002comparative,li2004comparative,8242675,9774526,liao2022cfs,liao2022tuning}. Furthermore, in the era of Deep Learning (DL), feature selection has been recently used to analyze deep neural networks by providing interpretations and explanations~\cite{lu2018deeppink}.

The last decades have seen the fast development of DL, it becomes more and more appealing to exploit concepts from deep learning to design and ameliorate feature selection algorithms. Naturally, DL-based feature selection algorithms are born with computational efficiency due to parallel computing on Graphics Processing Units and high scalability to large-scale datasets. Considering that selection itself is actually an  NP-hard discrete optimization problem, DL-based FS methods often aim to approximate the optimal solution and therefore relax the feature selection problem into searching a feature mask\footnote{In some literature feature mask is also known as a vector of feature importance scores and we use both terms interchangeably in this paper.} that maximizes the performance of the jointly trained learning algorithm (typically implemented by a neural network) with some constraints on the feature mask~\cite{gui2019afs,abid2019concrete,han2018autoencoder,borisov2019cancelout,roy2015feature,li2016deep,liao2021feature,mirzaei2020deep}. In this case, each entry of the feature mask denotes the importance of the corresponding candidate feature. Feature selection is accordingly performed after training by identifying the $k$ features with the greatest importance scores. Thereby, one of the major research trends is to design novel learning objectives and special architectures to regularize feature masks during training. According to us, broadly speaking, there are three main strategies for DL-based FS methods, namely sparsity-based, stochastic-based and attention-based feature selection algorithms. Sparsity-based methods such as~\cite{li2016deep,borisov2019cancelout,han2018autoencoder,wu2021fractal} force most entries of the feature mask to be small or exactly zeros during training so that important features have significantly larger weights than irrelevant or noisy features. Stochastic-based methods~\cite{abid2019concrete,trelin2020binary,dona2021differentiable} use reparameterization and relaxation of discrete distributions to explore more different feature combinations or importance scores during training. Attention-based approaches~\cite{gui2019afs,liao2021feature} construct a direct mapping between input data and feature masks and therefore expect that neural networks can automatically learn more reasonable and suitable feature masks dependent of the input data. It should be additionally noted that many existing methods often combine the above mentioned three strategies to construct a more complex and powerful algorithm for feature selection.

Although the aforementioned DL-based FS approaches have shown various design paradigms, one interesting common characteristic is that the candidate features with large importance scores are dominant for the joint learning network during training, while the learning objective does not consider the features with small scores. Based on this observation, we raise a twofold concern. Firstly, some relevant features might be undesirably down-weighted and ignored during training as stated in~\cite{zheng2021feature}. For example, the feature mask at a certain training step may coincidentally assign large scores to \emph{partial} strongly relevant features and some weakly relevant or even unimportant features. Meanwhile, the rest relevant features are assigned with small scores only. Despite this suboptimal feature mask, the training loss can be still small. This is effortful for algorithms to explore other feature masks and the training can stop in some local minima. As a result, some important features are unfortunately neglected due to their falsely assigned small importance scores. Secondly, it is intuitive that a good prediction performance (e.g. high accuracy for classification) indicates good candidate features. However, we are asking: Does poor prediction performance (e.g. low accuracy) indicate unimportant features? Now we consider feature selection towards a binary classification problem as an example. If the selected features result in the poorest prediction performance with 0\% accuracy, the selected features actually can optimally discriminate both classes by simply inverting the predicted class labels. Thereby, we argue that unimportant features should be the features with \emph{uncertain} predictive ability instead of poor prediction performance. Unfortunately, to the best of our knowledge, no prior work has considered this property during designing FS algorithms.

The twofold concern above motivates us to rethink the feature selection problem by considering those features with small importance scores during training. That is to say, during training, the feature mask should assign small weights to as few relevant features as possible, and meanwhile, the features with small scores should have uncertain predictive capability. To this end, our paper proposes a feature selection framework based on a novel complementary feature mask. In addition to an ordinary feature mask in many FS algorithms, we specifically define a complementary feature mask and use it to improve the feature selection performance. In total, the main contributions of this work are summarized as follows:
\begin{itemize}
	\item We have proposed a novel complementary feature mask method that considers features assigned with small or incorrect importance scores during training;
	\item A generic multi-task feature selection framework leveraging the complementary feature mask is presented which can be considered as a paradigm for designing new feature selection algorithms;
	\item We provide an implementation of the new framework and have conducted extensive experiments to show its effectiveness and superiority.
\end{itemize}

\section{Background}
\label{sec:background}
In this paper, we use the following notations. Let the input data with the candidate features be as $\mathcal{X} = \{\bm{x}_1, \bm{x}_2, \dots, \bm{x}_N\}$ with $N$ samples, each sample is denoted as a column vector $\bm{x}_i\in\mathbb{R}^D$, meaning that we have $D$ candidate features. In the following, the input data are also denoted as a matrix $X=[\bm{x}_1, \bm{x}_2, \dots, \bm{x}_N]^\top\in\mathbb{R}^{N\times D}$ and we use both notations interchangeably for better readability. Without loss of generality, we suppose the the ground truth label for each sample is a scalar in a supervised learning setup for classification, i.e. $\bm{y} = [y_1, y_2, \dots, y_N]^\top$ with $y_i\in\{1,2,\dots,C\}$ where $C$ is the number of classes. Typically, we use one-hot-coding representations for the class labels and $\bm{y}$ can be thus formulated as $Y=[\bm{e}_1, \bm{e}_2, \dots, \bm{e}_N]^\top\in\mathbb{R}^{N\times C}$ where $\bm{e}_i\in\{0, 1\}^C$ and the only non-zero entry is the affiliation of the corresponding sample $\bm{x}_i$; i.e. the $y_i$-th entry of $\bm{e}_i$ is one while the other entries are zeros. 

\begin{figure}
	\centering
	\begin{tikzpicture}[scale=0.75]
		\node (x) at(-3, 0) {$X$};
		\node (m) at(3, -1.5) {$\bm{m}$};
		\node (odot) at (3, 0) {$\odot$};
		\node (y) at(7.5, 0) {$\hat{Y}$};
		
		\draw[thick, draw=black!55, rounded corners=1mm,fill=orange!40,fill opacity=0.7] (4, -1.25) -- (4, 1.25) -- (6, 0.6) -- (6, -0.6) -- cycle;
		\draw[thick, dashed, draw=black!55, rounded corners=1mm,fill=blue!40,fill opacity=0.7] (2, -0.5) -- (2, -2.5) -- (-1, -2.5) -- (-1, -0.5) -- cycle;
		
		\draw[thick, black!55, dashed, ->] (x) -- (-2, 0) -- (-2, -1.5) -- (-1, -1.5);
		\draw[thick, black!55, dashed, ->] (2, -1.5) -- (m);
		\draw[thick, black!55, ->] (m) -- (odot);
		\draw[thick, black!55, ->] (x) -- (odot);
		\draw[thick, black!55, ->] (odot) -- (4, 0);
		\draw[thick, black!55, ->] (6, 0) -- (y);
		\draw[fill=black!55, draw=black!55] (-2, 0) circle(0.1);
		
		\node (g) at(5, 0) {\large$g_{\bm{\Theta}_n}(\cdot)$};
		\node (f) at(0.5, -1.5) {\large$f_{\bm{\Theta}_m}(\cdot)$};
	\end{tikzpicture}
	\caption{A typical DL-based feature selection framework. It jointly learns the feature mask $\bm{m}$ and a learning network $g_{\bm{\Theta}_n}(\cdot)$. For some methods, $\bm{m}$ is dependent of the input data $X$ and this path is illustrated in a dashed line.}
	\label{fig:dl-fs}
\end{figure}
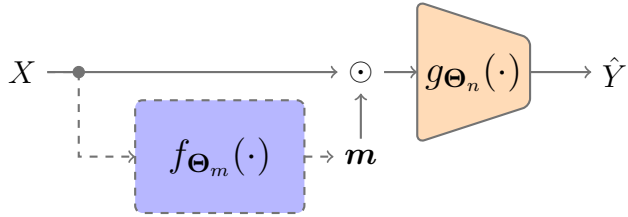

Broadly speaking, according to us, most DL-based feature selection algorithms follow the structure shown in Figure~\ref{fig:dl-fs}. Put simply, they simultaneously learn a feature mask $\bm{m}=[m_1, m_2,\dots, m_D]^\top\in\mathbb{R}^D$ and a neural network $g_{\bm{\Theta}_n}(\cdot)$ that maximizes a prediction performance with carefully designed constraints on $\bm{m}$ and/or special generation process of $\bm{m}$. As introduced before, each entry $m_i$ of $\bm{m}$ indicates the importance of the corresponding $i$-th candidate feature and $m_i$ is therefore often non-negative. For example, some studies such as~\cite{trelin2020binary} require the values to be in the range of $[0, 1]$, while some other studies such as~\cite{liao2021feature,gui2019afs} restrict the values to be in $(0, 1)$. 

In Figure~\ref{fig:dl-fs}, the dashed line illustrates the path from input $X$ to the feature mask $\bm{m}$. This is because some prior works such as~\cite{li2016deep,borisov2019cancelout} assume that $\bm{m}$ is independent of $X$ during the forward propagation. On the contrary, some recent methods such as~\cite{gui2019afs,liao2021feature} leverage attention mechanism and $\bm{m}$ is thus directly dependent of the input data which is calculated as $\bm{m} = f_{\bm{\Theta}_m}(X)$. During training, for many methods, additional regularization $\mathcal{R}(\cdot)$ must be applied to the feature mask $\bm{m}$ to guarantee certain desired properties like sparsity in $\bm{m}$. In total, the learning objective of DL-based feature selection methods can be formulated as
\begin{equation}
	\label{eq:dl-fs-objective}
	\argmin_{\bm{m}, \bm{\Theta}} \mathcal{L}(g(\bm{x}\odot\bm{m}), y) + \lambda\cdot \mathcal{R}(\bm{m}),
\end{equation}
where $\bm{\Theta}=\{\bm{\Theta}_n, \bm{\Theta}_m\}$ and we omit the subscript for $g_{\bm{\Theta}_n}(\cdot)$ for simplicity. In the learning objective, the first term is a loss function towards the given learning task (e.g. classification) and $\lambda$ is a weighting factor balancing the two terms. One popular regularization is to force the sparsity of $\bm{m}$ during training as in~\cite{li2016deep}. Some other methods such as~\cite{abid2019concrete,trelin2020binary} do not pose sparsity penalty on $\bm{m}$ but generates $\bm{m}$ from a relaxation of discrete distributions and it is expected to explore more different feature subset combinations during training. In should be additionally mentioned that the common regularization (e.g. $\ell_2$ weight decay) on the trainable parameters of neural networks can be added to the learning objective to avoid overfitting, but such regularization is not directly related to the FS performance and we omit it for simplicity. Finally, after training, we can select the top-$k$ features by comparing the importance scores in $\bm{m}$.

\section{Related Works}
\label{sec:related-work}
Over the last two decades, a considerable literature has grown up around the theme of feature selection due to its importance and wide applications. Broadly speaking, as in~\cite{guyon2003introduction,chandrashekar2014survey,li2017feature}, it is generally accepted that feature selection methods can be categorized into three groups: wrapper methods, filter methods and embedded methods. Wrapper methods evaluate all possible combinations of features for a given predictive model and thus not generate any feature masks. Theoretically, wrapper methods can yield the optimal feature subset with a fixed size with respect to a given learning algorithm (e.g. a given classifier). However, it is not feasible if we encounter data with high dimensions~\cite{li2017feature}. In contrast, filter methods do not rely on a specific predictive model and assess features according to a given criteria such as correlation~\cite{guyon2003introduction} and mutual information~\cite{peng2005feature}. Although filter methods possess a higher computational efficiency than wrapper methods due to the absence of training predictive models, complex relations between features can be therefore difficult to detect based on simple criteria and this often results in suboptimal solutions. Embedded methods, however, can be considered as a combination of wrapper and filter methods~\cite{li2017feature}. They integrate learning algorithms into feature selection process; i.e. embedded methods simultaneously learn to weight features and train a predictive model on the weighted features.

Based on the categorization above, most DL-based feature selection methods can be considered as embedded methods because they combine learning feature masks and training predictive models together. Despite the recent boom of DL-based feature selection methods, to the best of our knowledge, nearly no prior studies have paid attention to importance score order or leveraging the features with small scores during training. The most related works to us can be the feature selection algorithms partially supported by unselected features. \cite{caruana2003benefitting} first noticed the advantage of discarded features. In particular, they proposed a concept that the unselected features can be useful for classification tasks in a multi-task learning framework. Afterwards, \cite{taylor2016learning} extended this idea to further improve the performance of predictive models. Nonetheless, both studies did not provide new feature selection methods but aimed to enhance the downstream learning performance. A recent work~\cite{zheng2021feature} leveraged the unselected features and design a minimax optimization problem between selected and unselected features. However, it restricted with linear models and has complex training procedures, which make it difficult to extend to other methods.
 
\section{Methodology}
\label{sec:method}
The key idea is to introduce a Complementary Feature Mask (CFM) $\tilde{\bm{m}}$ to the original feature mask $\bm{m}$ so that large losses can be expected if a few candidate features are assigned with incorrect importance scores during training. This means that the jointly trained neural network can exploit the features that are occasionally neglected or incorrectly weighted by $\bm{m}$ but thus captured by $\tilde{\bm{m}}$.
\begin{definition}[Complementary Feature Mask]
	$\tilde{\bm{m}}$ is the complementary feature mask of a given feature mask $\bm{m}$ if the ranking of elements in $\tilde{\bm{m}}$ is opposite to that of $\bm{m}$.
\end{definition}
In order to force the candidate features with incorrect weighting to have poor predictive capability, it is natural to design a multi-task learning framework to learn $\bm{m}$ by simultaneously considering $\tilde{\bm{m}}$. The \emph{main} learning objective is the same as Equation~\ref{eq:dl-fs-objective}. That is to say, the entry in $\bm{m}$ with a greater value indicates a more important feature. On the other hand, as an auxiliary learning task, $\tilde{\bm{m}}$ assigns complementary importance scores to all features and forces $\bm{x}\odot\tilde{\bm{m}}$ to have poorer predictive ability and this side is called \emph{complementary}. Thereby, the overall multi-task learning objective is defined as
\begin{equation}
	\label{eq:new-objective}
	\argmin_{\bm{m}, \bm\Theta} \mathcal{L}(g(\bm{x}\odot\bm{m}), y) + \lambda\cdot\mathcal{R}(\bm{m}) + \gamma\cdot\mathcal{L}_{c}(g(\bm{x}\odot\tilde{\bm{m}})),
\end{equation}
where the first two terms are exactly the original learning objective for most DL-based FS algorithms defined in Equation~\ref{eq:dl-fs-objective}, while $\mathcal{L}_{c}(\cdot)$ is a special learning objective for the proposed novel CFM that measures how poor the predictive ability of $\bm{x}\odot\tilde{\bm{m}}$ is. Put simply, a more uncertain prediction in the complementary path corresponds to a smaller loss of $\mathcal{L}_{c}(\cdot)$, which is not inclusive by the main path. Thereby, the minimization of Equation~\ref{eq:new-objective} equals to searching for such features that can lead to better prediction performance in the main path, while forcing that the features with complementary importance scores only pose uncertain prediction for the given task. As a result, really important and representative features can be scored in a correct order and thus selected. 

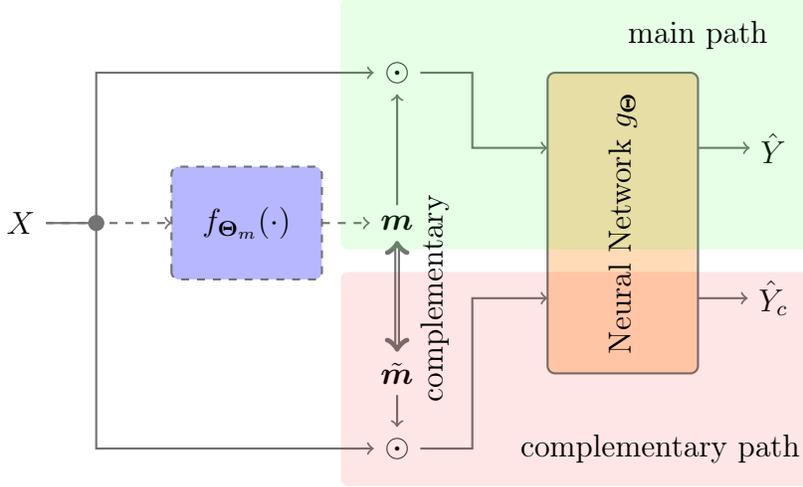
\begin{figure}
	\centering
	\begin{tikzpicture}[scale=1.]
		\node (x) at(0, 0) {$X$};
		\node (odot1) at(5, 2) {$\odot$};
		\node (odot2) at(5, -3) {$\odot$};
		\node (m) at(5, 0) {$\bm{m}$};
		\node (m-neg) at(5, -2) {$\tilde{\bm{m}}$};
		\node (pos) at(10, 1) {$\hat{Y}$};
		\node (neg) at(10, -1) {$\hat{Y}_{c}$};
		
		\draw[thick, black!55, dashed, rounded corners=1mm,fill=blue!40,fill opacity=0.7] (2, -0.75) -- (2, 0.75) -- (4, 0.75) -- (4, -0.75) -- cycle;
		\draw[thick, black!55, rounded corners=1mm,fill=orange!40,fill opacity=0.7] (7, 2) -- (7, -2) -- (9, -2) -- (9, 2) -- cycle;
		
		\draw[thick, black!55, dashed, ->] (x) -- (2, 0);
		\draw[thick, black!55] (x) -- (1, 0);
		\draw[thick, black!55, dashed, ->] (4, 0) -- (m);
		\draw[thick, black!55, ->] (m) -- (odot1);
		\draw[thick, black!55, ->] (m-neg) -- (odot2);
		\draw[thick, black!55, ->] (1, 0) -- (1, 2) -- (odot1);
		\draw[thick, black!55, ->] (1, 0) -- (1, -3) -- (odot2);
		\draw[thick, black!55, ->] (odot1) -- (6, 2) -- (6, 1) -- (7, 1);
		\draw[thick, black!55, ->] (odot2) -- (6, -3) -- (6, -1) -- (7, -1);
		\draw[thick, black!55, ->] (9, 1) -- (pos);
		\draw[thick, black!55, ->] (9, -1) -- (neg);
		\draw[thick, black!55, <->, double] (m) -- (m-neg);
		\draw[fill=black!55, draw=black!55] (1, 0) circle(0.1);
		
		\fill [rounded corners=1mm, green, fill opacity=0.1] (4.25, 3) rectangle (10.5, -0.35);
		\fill [rounded corners=1mm, red, fill opacity=0.1] (4.25, -3.5) rectangle (10.5, -0.65);
		
		\node [rotate=90] (mm) at(5.5, -1) {complementary};
		
		\node (pos-lb) at(9, 2.5) {main path};
		\node (neg-lb) at(8.5, -3) {complementary path};
		
		\node [rotate=90] (g) at(8, 0) {Neural Network $g_{\bm{\Theta}}$};
		\node (f) at(3, 0) {$f_{\bm{\Theta}_m}(\cdot)$};
	\end{tikzpicture}
	\caption{The proposed generic framework for feature selection using complementary feature mask for an auxiliary learning task. }
	\label{fig:fscr}
\end{figure}

To apply the novel CFM into existing DL-based approaches, it is required to slightly adapt the original feature selection framework in Figure~\ref{fig:dl-fs} to a multi-task structure as shown in Figure~\ref{fig:fscr}. In addition to the original structure (main path in green), we additionally feed $\bm{x}\odot\tilde{\bm{m}}$ to the same neural network to obtain the complementary prediction $\hat{y}_{c}$ and this prediction should be uncertain during training. Additionally, it is to notice that the calculation of $\tilde{\bm{m}}$ and $\mathcal{L}_c$ is generic and the concrete design depends on a given learning task and prior information. In the following section, we present our implementation of this framework.

\section{Implementation}
\label{sec:implementation}
In the previous section, we have introduced the key idea of the proposed multi-task feature selection framework using the novel complementary feature mask. Accordingly, this section demonstrates how to implement the novel CFM method to perform feature selection.

\subsection{Complementary Feature Mask}
We start with the feature mask $\bm{m}$. Inspired by the recent work~\cite{liao2021feature} that has achieved state-of-the-art feature selection performance, $\bm{m}$ is defined as:
\begin{equation}
	\label{eq:fm}
	\bm{m} = \mathit{softmax}\Big(\frac{1}{B}\sum_{i=1}^{B}\big(W_2\tanh(W_1\bm{x}_i + \bm{b}_1) + \bm{b}_2\big)\Big),
\end{equation}
where $B$ denotes the minibatch size during training and this equation defines the mapping between the minibatch input data $X$ and feature mask $\bm{m}$ as $\bm{m} = f_{\bm{\Theta}_m}(X)$. Thereby, $\bm{\Theta}_m = \{W_1, W_2, \bm{b}_1, \bm{b}_2\}$ are trainable parameters. In order to guarantee that the values of both $\bm{m}$ and $\tilde{\bm{m}}$ are of similar scale during training, the complementary feature mask $\tilde{\bm{m}}$ is defined as
\begin{equation}
	\label{eq:coupled-mask}
	\tilde{\bm{m}} = \mathit{softmax}\Big(-\frac{1}{B}\sum_{i=1}^{B}\big(W_2\tanh(W_1\bm{x}_i + \bm{b}_1) + \bm{b}_2\big)\Big).
\end{equation}
We can clearly see that the ranking of the elements in $\tilde{\bm{m}}$ is opposite to that of $\bm{m}$ and the values of both $\bm{m}$ and $\tilde{\bm{m}}$ are of similar scale.

\subsection{Architecture}
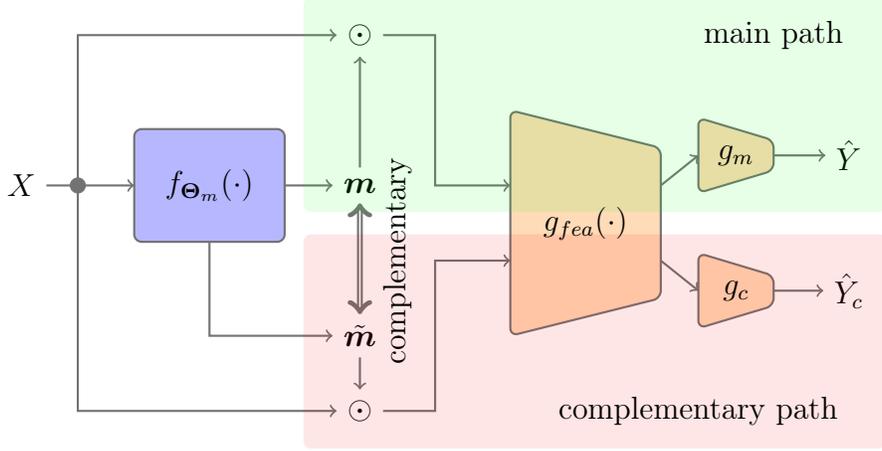
\begin{figure}
	\centering
	\begin{tikzpicture}[scale=1]
		\node (x) at(0.5, 0) {$X$};
		\node (odot1) at(5, 2) {$\odot$};
		\node (odot2) at(5, -3) {$\odot$};
		\node (m) at(5, 0) {$\bm{m}$};
		\node (m-neg) at(5, -2) {$\tilde{\bm{m}}$};
		\node (pos) at(11.5, 0.4) {$\hat{Y}$};
		\node (neg) at(11.5, -1.4) {$\hat{Y}_{c}$};
		
		\draw[thick, black!55, rounded corners=1mm,fill=blue!40,fill opacity=0.7] (2, -0.75) -- (2, 0.75) -- (4, 0.75) -- (4, -0.75) -- cycle;
		\draw[thick, black!55, ->] (3, -0.75) -- (3, -2) -- (m-neg);
		\draw[thick, black!55, rounded corners=1mm,fill=orange!40,fill opacity=0.7] (7, 1) -- (7, -2) -- (9, -1.5) -- (9, 0.5) -- cycle;
		
		\draw[thick, black!55, rounded corners=1mm,fill=orange!40,fill opacity=0.7] (9.5, 0.9) -- (9.5, -0.1) -- (10.5, 0.2) -- (10.5, 0.6) -- cycle;
		\draw[thick, black!55, rounded corners=1mm,fill=orange!40,fill opacity=0.7] (9.5, -0.9) -- (9.5, -1.9) -- (10.5, -1.6) -- (10.5, -1.2) -- cycle;
		
		\draw[thick, black!55, ->] (x) -- (2, 0);
		\draw[thick, black!55, ->] (4, 0) -- (m);
		\draw[thick, black!55, ->] (m) -- (odot1);
		\draw[thick, black!55, ->] (m-neg) -- (odot2);
		\draw[thick, black!55, ->] (1.25, 0) -- (1.25, 2) -- (odot1);
		\draw[thick, black!55, ->] (1.25, 0) -- (1.25, -3) -- (odot2);
		\draw[thick, black!55, ->] (odot1) -- (6, 2) -- (6, 0) -- (7, 0);
		\draw[thick, black!55, ->] (odot2) -- (6, -3) -- (6, -1) -- (7, -1);
		\draw[thick, black!55, ->] (9, 0) -- (9.5, 0.4);
		\draw[thick, black!55, ->] (9, -1) -- (9.5, -1.4);
		\draw[thick, black!55, ->] (10.5, 0.4) -- (pos);
		\draw[thick, black!55, ->] (10.5, -1.4) -- (neg);
		\draw[thick, black!55, <->, double] (m) -- (m-neg);
		\draw[fill=black!55, draw=black!55] (1.25, 0) circle(0.1);
		
		\fill [rounded corners=1mm, green, fill opacity=0.1] (4.25, 2.5) rectangle (12., -0.35);
		\fill [rounded corners=1mm, red, fill opacity=0.1] (4.25, -3.5) rectangle (12., -0.65);
		
		\node [rotate=90] (mm) at(5.5, -1) {complementary};
		
		\node (pos-lb) at(10.5, 2.) {main path};
		\node (neg-lb) at(9.5, -3) {complementary path};
		
		\node (pos-out) at(10, 0.4) {$g_{m}$};
		\node (neg-out) at(10, -1.4) {$g_{c}$};
		
		\node (g) at(8, -0.5) {$g_{fea}(\cdot)$};
		\node (f) at(3, 0) {$f_{\bm{\Theta}_m}(\cdot)$};
	\end{tikzpicture}
	\caption{The architecture of our implementation for CFM.}
	\label{fig:nn}
\end{figure}
Figure~\ref{fig:nn} shows the general architecture of our implementation. The most conspicuous design is that we use a multi-output architecture to realize $g_{\bm{\Theta}_n}(\cdot)$. This design is inspired by feature learning~\cite{schlachter2019deep} and multi-task learning~\cite{caruana2003benefitting,caruana1997multitask}. The network can be therefore understood as being composed of several feature extraction layers and two separate classification layers. Accordingly, $g_{fea}$ consists of all layers from the first layer to the penultimate layer acting as a feature extractor for the input data, while it is followed by two independent dense layers, one for the main path $g_m$ and one for the complementary path $g_c$. This design ensures that the network maps $\bm{x}\odot\bm{m}$ and $\bm{x}\odot \tilde{\bm{m}}$ into the same feature space. This makes more sense for the requirement that the original feature mask should select features leading to better performance than those selected by the complementary feature mask. 

\begin{figure}
	\centering
	\begin{tikzpicture}[scale=.65]
		\node (x1) at(-1.5, 1) {$X\odot \bm{m}$};
		\node (x2) at(-1.5, -1) {$X\odot \tilde{\bm{m}}$};
		
		\draw[thick, draw=black!55, rounded corners=1mm,fill=orange!40,fill opacity=0.7] (0.2, -3.2)--(10.8, -2.2)--(10.8, 2.2)--(0.2, 3.2)--cycle;
		\node (g) at(1, 2.5) {$g_{fea}$};
		
		\draw[thick, draw=black!55, rounded corners=1mm,fill=yellow!40,fill opacity=0.7] (1, -2)--(2, -2)--(2, 2)--(1, 2)--cycle;
		\node [rotate=90] (d1) at(1.5, 0) {dense-128};
		
		\draw[thick, draw=black!55, rounded corners=1mm,fill=blue!30,fill opacity=0.7] (3, -2)--(4, -2)--(4, 2)--(3, 2)--cycle;
		\node [rotate=90] (d1) at(3.5, 0) {LeakyReLU};
		
		\draw[thick, draw=black!55, rounded corners=1mm,fill=yellow!40,fill opacity=0.7] (5, -2)--(6, -2)--(6, 2)--(5, 2)--cycle;
		\node [rotate=90] (d1) at(5.5, 0) {dense-64};
		
		\draw[thick, draw=black!55, rounded corners=1mm,fill=blue!30,fill opacity=0.7] (7, -2)--(8, -2)--(8, 2)--(7, 2)--cycle;
		\node [rotate=90] (d1) at(7.5, 0) {LeakyReLU};
		
		\draw[thick, draw=black!55, rounded corners=1mm,fill=lightgray!20,fill opacity=0.7] (9, -2)--(10, -2)--(10, 2)--(9, 2)--cycle;
		\node [rotate=90] (d1) at(9.5, 0) {Dropout};
		
		\draw[thick, draw=black!55, rounded corners=1mm,fill=green!20,fill opacity=0.7] (11.5, 0.5)--(12.5, 0.5)--(12.5, 4.5)--(11.5, 4.5)--cycle;
		\node [rotate=90] (d1) at(12., 2.5) {dense-$C$};
		\node (gm) at(11, 4.) {$g_{m}$};
		\node (y1) at(13.5, 2.5) {$\hat{Y}$};
		
		\draw[thick, draw=black!55, rounded corners=1mm,fill=pink!40,fill opacity=0.7] (11.5, -4.5)--(12.5, -4.5)--(12.5, -0.5)--(11.5, -0.5)--cycle;
		\node [rotate=90] (d1) at(12., -2.5) {dense-$C$};
		\node (gc) at(11, -4.) {$g_{c}$};
		\node (y2) at(13.5, -2.5) {$\hat{Y}_c$};
		
		\draw[thick, black!55, ->] (x1) -- (1, 1);
		\draw[thick, black!55, ->] (x2) -- (1, -1);
		\draw[thick, black!55, ->] (2, 0) -- (3, 0);
		\draw[thick, black!55, ->] (4, 0) -- (5, 0);
		\draw[thick, black!55, ->] (6, 0) -- (7, 0);
		\draw[thick, black!55, ->] (8, 0) -- (9, 0);
		
		\draw[thick, black!55, ->] (10, 1) -- (11.5, 2.5);
		\draw[thick, black!55, ->] (10, -1) -- (11.5, -2.5);
		\draw[thick, black!55, ->] (12.5, 2.5) -- (y1);
		\draw[thick, black!55, ->] (12.5, -2.5) -- (y2);
	\end{tikzpicture}
	\caption{The architecture of the learning network used for the proposed CFM.}
	\label{fig:cfm}
\end{figure}
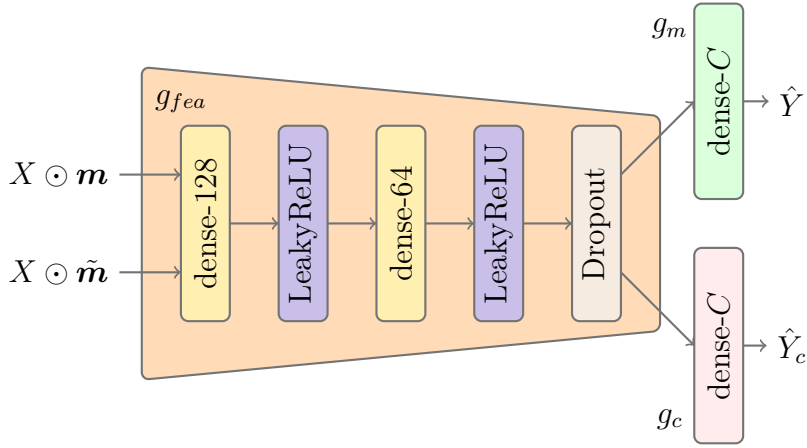

The learning network mentioned above was implemented as in Figure~\ref{fig:cfm}. The feature extraction network $g_{fea}$ was composed of two hidden dense layers with $128$ and $64$ neurons respectively, a dropout layer with the rate of $0.3$ following the two dense layers, and one output dense layer for classification. The activation function for both hidden layers was LeakyReLU~\cite{maas2013rectifier} with $\alpha=0.02$. For CFM, as shown in Figure~\ref{fig:cfm}, we used two separate dense layers to concatenate to the dropout layer to enable the multi-task architecture. All output layers use the activation of $\mathit{softmax}$ due to the classification task. It should be noted that in Figure~\ref{fig:cfm}, we omit the partial network for generating $\bm{m}$ for simplicity.

\subsection{Learning Objective}
Similar to~\cite{liao2021feature}, the proposed CFM method does not possess any additional regularization on $\bm{m}$. Therefore, the learning objective is simplified to
\begin{equation}
	\label{eq:cfm-objective}
	\argmin_{\bm{m}, \bm\Theta} \mathcal{L}(g_m(g_{fea}(\bm{x}\odot\bm{m})), y)  + \gamma\cdot\mathcal{L}_{c}(g_c(g_{fea}(\bm{x}\odot\tilde{\bm{m}}))).
\end{equation}
To demonstrate the proposed framework, we first restrict our experiments to supervised classification for simplicity. The choice of $\mathcal{L}$ is intuitive as we can simply use a categorical cross entropy loss. On the other hand, the design for $\mathcal{L}_c$ is not easy because there are many ways to achieve a poor prediction performance, but not each poor solution of the complementary path is related to a good choice of the feature mask $\bm{m}$ and its corresponding complementary feature mask $\tilde{\bm{m}}$. We suggest that it might be not reliable to directly maximize the cross entropy loss between the prediction $\hat{\bm{y}}_{c}$ and the ground truth $\bm{y}$ to achieve a poor prediction performance because it can lead to trivial and biased solution. For example, if the complementary path always outputs a certain but fixed class, this is a poor prediction but it does not improve the feature selection of $\bm{m}$. In contrast, as mentioned before, a poor predictive capability means uncertain predictions for $\bm{x}\odot\tilde{\bm{m}}$; i.e. a given input sample can be randomly classified to any one of the known classes. Thereby, this work defines $\mathcal{L}_{c}$ as
\begin{equation}
	\label{eq:loss-comp}
	\mathcal{L}_{c} = - \frac{1}{N}\sum_{i=1}^{N} \bm{e}_{r,i}^T\cdot \log\Big(g_{c}(g_{fea}(\bm{x}_i\odot\tilde{\bm{m}}))\Big),
\end{equation}
where $\bm{e}_{r,i}$ is the one-hot coding of a random class label for a sample $\bm{x}_i$. Specifically, the random class label is sampled from a discrete uniform distribution $\mathcal{U}(1, C)$. As can be seen, minimizing the loss defined in Equation~\ref{eq:loss-comp} maximizes the uncertainty of the predictions, because the samples from the same class can be predicted to different classes. 
\section{Experiments}
\label{sec:exp}
This section evaluates the proposed FS framework based on the novel complementary feature mask. Specifically, we conducted experiments on six popular datasets of different types including images, texts and speech signals. In addition, we compared our method to one of the most recent state-of-the-art DL-based FS approaches to justify the selection quality of our framework. 
\subsection{Setup}
\label{subsec:setup}
In general, we follow the similar experimental settings to the prior works~\cite{gui2019afs,abid2019concrete,liao2021feature}; i.e. we trained feature selection algorithms on the given datasets and selected $k$ features according the learned feature mask. Subsequently, we evaluated these $k$ features on a separate classifier. Finally, the resulting classification accuracy on the test set with the selected features indicated the selection quality. We repeated the experiments with five different random seeds and report the averaged results with deviation for objective analysis.

\subsubsection{Reference Method}
\label{subsubsec:reference-methods}
As comparison, we used the Feature Mask (FM) method~\cite{liao2021feature} as a reference method in our experiments. According to~\cite{liao2021feature}, the FM method achieved state-of-the-art feature selection performance on several benchmarking datasets in comparison with other recent DL-based approaches. Hence, FM can be considered as a convincing reference method and a challenging competitor to our approach. To have a fair comparison, both FM and our CFM had the same learning network (except that our method had one additional complementary path). The hyperparameter $\gamma$ of our method was chosen by simple grid search from $\{0.001, 0.01, 0.1, 1, 10, 100\}$ on a separate validation set (10\% of the training data).

\subsubsection{Datasets}
\label{subsubsec:datasets}
This paper uses six benchmarking datasets which are frequently used to evaluate feature selection methods, i.e. MNIST~\cite{lecun1998gradient}, Fashion-MNIST (fMNIST)~\cite{xiao2017fashion}, Isolet~\cite{fanty1991spoken}, PCMAC~\cite{Lang95}, madelon~\cite{guyon2007competitive} and gisette~\cite{guyon2007competitive}. An overview of the datasets are presented in Table~\ref{tab:datasets}.
\begin{table}
	\renewcommand{\arraystretch}{.85}
	\centering
	\caption{Details of the used datasets}
	\label{tab:datasets}
	\resizebox{.75\linewidth}{!}{
	\begin{tabular}{cccccc}
		\toprule
		& \bfseries Datasets & \bfseries \# Features & \bfseries  \# Train & \bfseries  \# Test & \bfseries  \# Classes\\
		\midrule
		1 & MNIST & 784 & 60000 & 10000 & 10\\
		2 & fMNIST & 784 & 60000 & 10000 & 10\\
		3 & Isolet & 617 & 1248 & 312 & 26 \\
		4 & PCMAC & 3289 & 1555 & 388 & 2\\
		5 & madelon & 500 & 2080 & 520 & 2 \\
		6 & gisette & 5000 & 5600 & 1400 & 2\\
		\bottomrule
	\end{tabular}
	}
\end{table}

\subsubsection{Downstream Classifiers}
\label{subsubsec:downstream-clf}
As mentioned before, we typically use the prediction results of downstream classifiers to indicate the quality of the selected features. In order to justify whether the selected features are robust to different classifiers, we use three different downstream classifiers to evaluate the selected features: i) Random Forest (RF)~\cite{breiman2001random}; ii) Extremely Randomized Tree (ERT)~\cite{geurts2006extremely}; iii) $k$-Nearest Neighbor (kNN)~\cite{guo2003knn}.

\subsection{Main Results}
\label{subsec:basic-exp}
\begin{figure}
	\centering
	\includegraphics[width=1\linewidth]{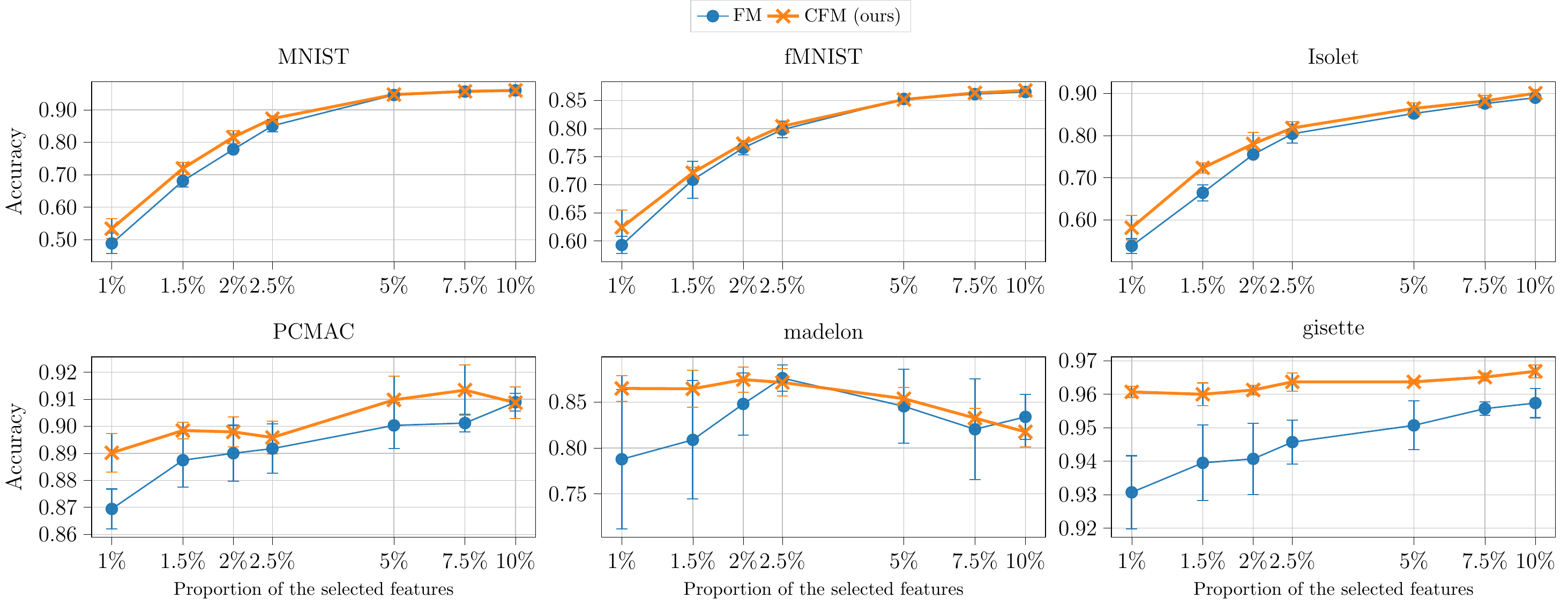}
	\caption{Classification accuracy on RF over different selected feature subset sizes.}
	\label{fig:accu_vs_k_rf}
\end{figure}
\begin{figure}
	\centering
	\includegraphics[width=1\linewidth]{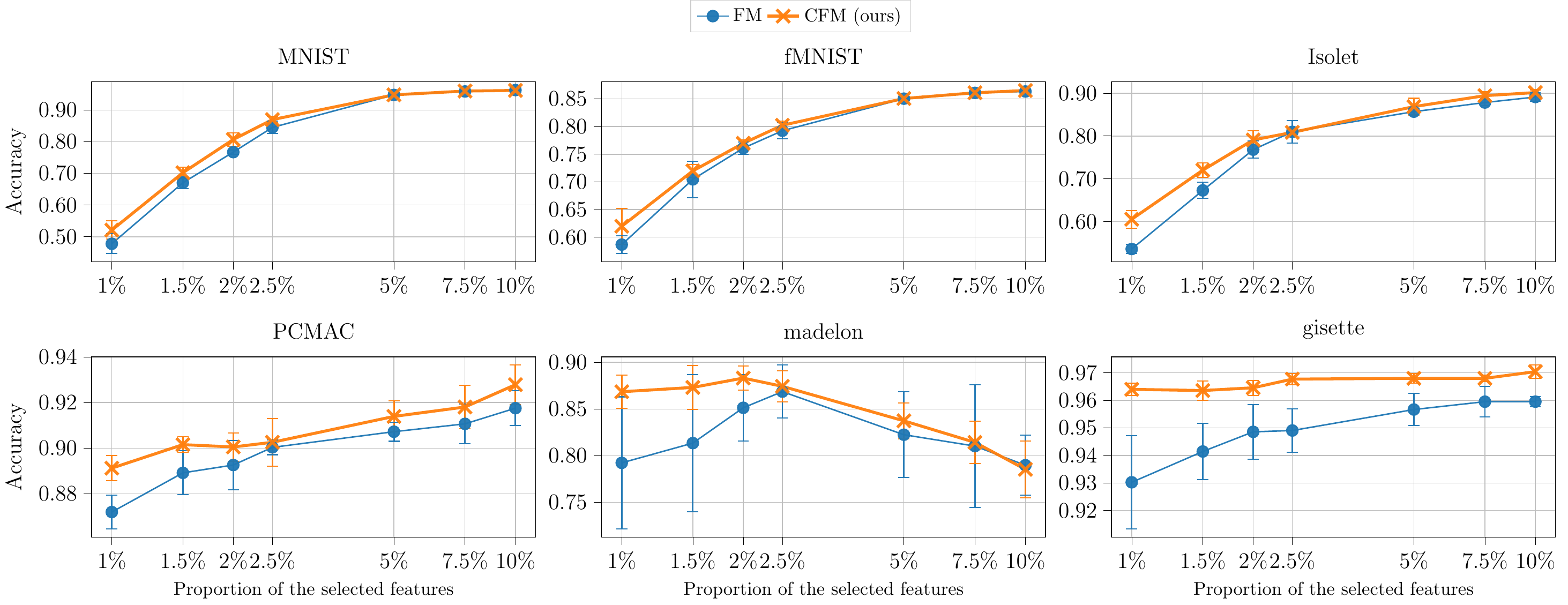}
	\caption{Classification accuracy on ERT over different selected feature subset sizes.}
	\label{fig:accu_vs_k_ert}
\end{figure}
\begin{figure}
	\centering
	\includegraphics[width=1\linewidth]{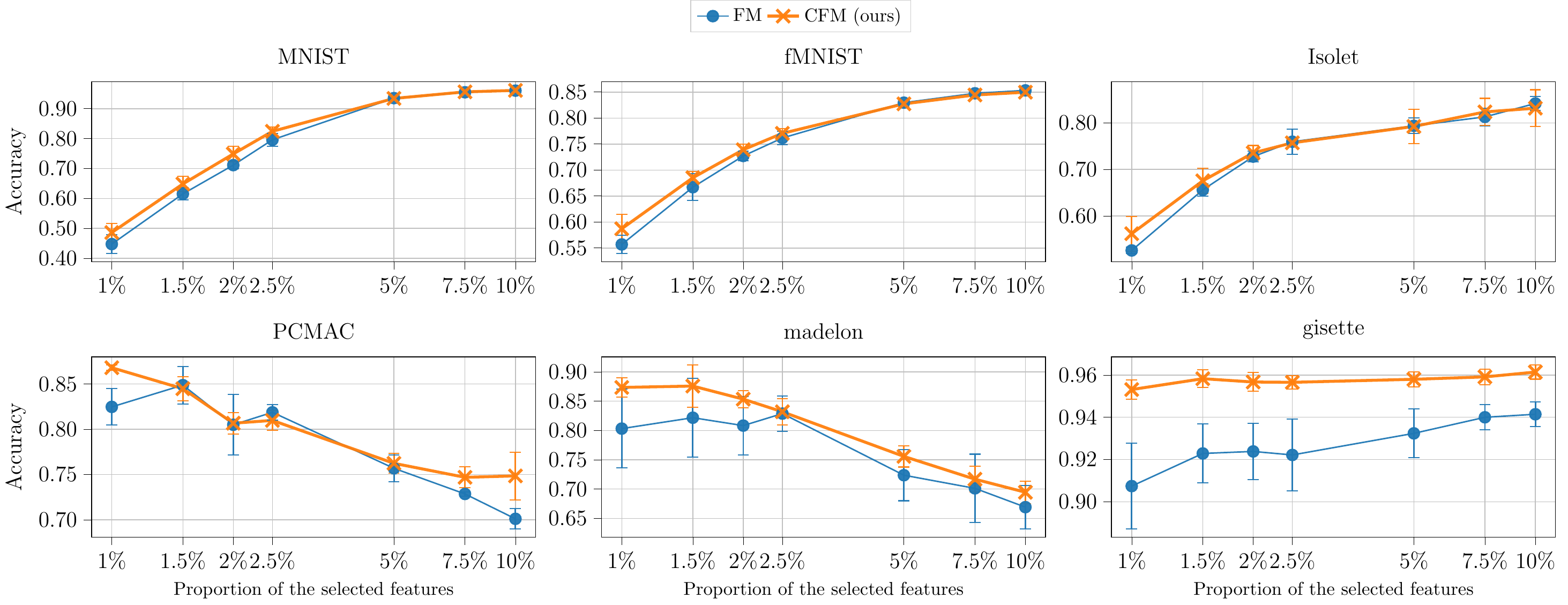}
	\caption{Classification accuracy on kNN over different selected feature subset sizes.}
	\label{fig:accu_vs_k_knn}
\end{figure}

The main experiments compared the proposed CFM framework with the FM method with respect to different feature subset sizes on the three aforementioned downstream classifiers. That is to say, the proportion of the selected features $\rho$ is defined as $\rho = \frac{k}{D}\times 100\%$, where $k$ denotes the number of selected features and $D$ denotes the number of raw features. In particular, in our experiments, we considered $\rho\in\{1\%, 1.5\%, 2\%, 2.5\%, 5\%, 7.5\%, 10\%\}$. For example, with $\rho=1\%$ for MNIST, we finally selected $0.01\times 784\approx8$ features (pixels) for training downstream classifiers.

Figure~\ref{fig:accu_vs_k_rf}, Figure~\ref{fig:accu_vs_k_ert} and Figure~\ref{fig:accu_vs_k_knn} present the experimental results on RF, ERT and kNN, respectively. Overall, our CFM method outperformed the state-of-the-art FM method in 109 out of 126 cases, corresponding to a significant better selection performance for different feature subset sizes and various downstream classifiers. This empirically confirms that the novel complementary path successfully regularized the learning procedure so that more representative and critical features were selected in a more reasonable order. 

One important observation is that our method resulted in notably smaller deviations in accuracy than those of the FM method for almost all cases. This can be easily seen especially for the datasets PCMAC, madelon and gisette. This experimental result suggests that our method has notably more stable selection results under different initialization seeds of neural networks, which remains, however, a challenge for many other DL-based feature selection approaches. One feasible reason is that the novel complementary feature mask forces the main path to learn feature importance in a correct order as much as possible. Better stability in selection results is specifically important for the use cases of understanding and analyzing high-dimensional data.

An additional observation is that our method generally performed extremely well for small feature subset sizes (i.e. small $\rho$), indicating that the learned feature mask really reflects the relative importance of each individual feature. On the contrary, the FM method without the regularization of complementary path cannot well maintain the correct importance order after training, although it can identify the overall important features.

Interestingly, the resulting accuracy for the madelon dataset decreased with increasing feature subset sizes for all three downstream classifiers as well as for both FM and our CFM methods. This is different from other datasets for most cases. The main reason might be that the madelon dataset was created with only 5 key features that directly correspond to the class label, 15 additional redundant features that were linear combinations of the 5 key features, and 480 distractor features. Thereby, feature subsets with large sizes indicate more misleading features are included and the downstream classification performance can be thereby negatively affected.

\subsection{Visualization of Learned Feature Masks}
\label{subsec:study-fea-mask}
\begin{figure}
	\centering
	\includegraphics[width=1\linewidth]{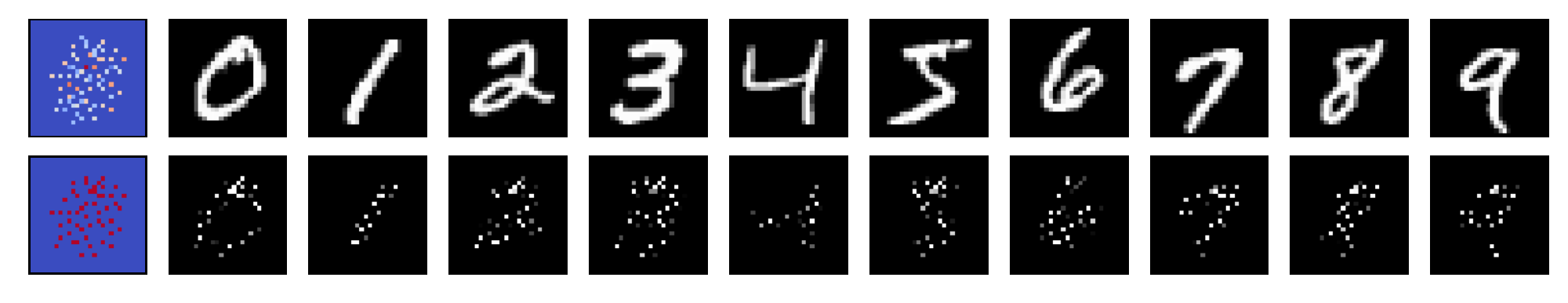}
	\caption{Visualization for the MNIST dataset: the learned feature mask (\emph{most top left}); the top-50 features based on the learned mask (\emph{most bottom left}); Exemplary images (\emph{the top row from second column}); the selected top-50 pixels based w.r.t. each image above (\emph{the bottom row from the second column}).}
	\label{fig:mnist_CFM_example}
\end{figure}
\begin{figure}
	\centering
	\includegraphics[width=1\linewidth]{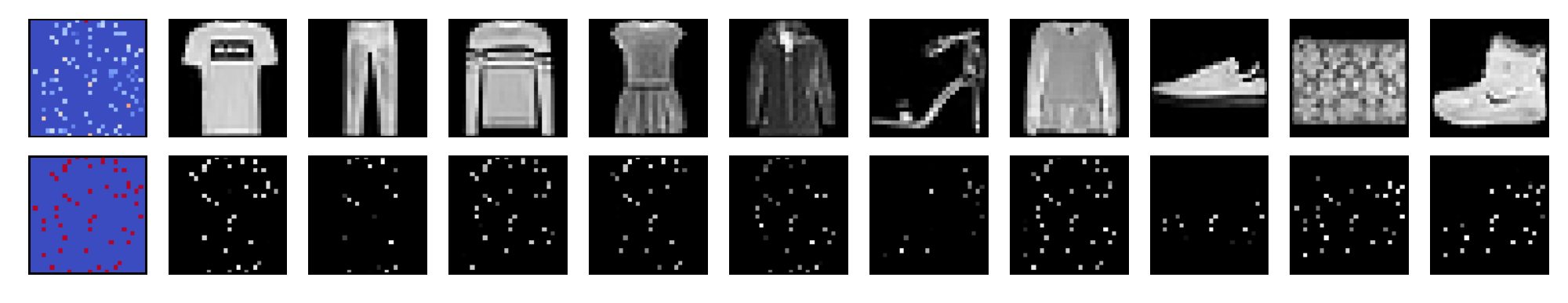}
	\caption{Visualization for the fMNIST dataset: the learned feature mask (\emph{most top left}); the top-50 features based on the learned mask (\emph{most bottom left}); Exemplary images (\emph{the top row from second column}); the selected top-50 pixels based w.r.t. each image above (\emph{the bottom row from the second column}).}
	\label{fig:fmnist_CFM_example}
\end{figure}
A challenge in feature selection is that we cannot directly understand how reasonable the selected features are due to the fact that exhaustive search is infeasible and we often do not have the ground truth of the optimal feature subsets. Nevertheless, fortunately, both MNIST and fMNIST datasets are image datasets. This enables us to visualize the learned feature masks and gain an intuitive understanding of the selected features.

Figure~\ref{fig:mnist_CFM_example} and Figure~\ref{fig:fmnist_CFM_example} present the results on MNIST and fMNIST, respectively. In each figure, the most top left image shows the learned feature masks normalized into 0 and 1 for better visualization (blue indicates smaller values, while red indicates larger values). The most bottom left image is the top-50 features (red pixels) selected based on the learned feature masks. In addition, the first row from the second column shows exemplary images randomly selected from the ten classes, while the second row from the second column shows the selected 50 pixels correspondingly. 

It can be seen that pixels in the center were mostly selected for the MNIST dataset, which makes sense, because digits are mostly located in the center of each image for the MNIST dataset. Furthermore, we can observe that even with the selected 50 pixels only, the digits are still recognizable, meaning that the critical structure of the input data is well maintained even after the selection procedure. The objects in the fMNIST dataset are typically large, so the learned most critical features are located more uniformly than those of MNIST. From the exemplary results of the fMNIST dataset, we can also observe that most critical and discriminative pixels of clothes, shoes and bags were maintained.

\subsection{The Design of Feature Masks}
\label{subsec:choice-of-fea-mask}
\begin{figure}
	\centering
	\begin{subfigure}{.5\textwidth}
		\centering
		\includegraphics[width=1\linewidth]{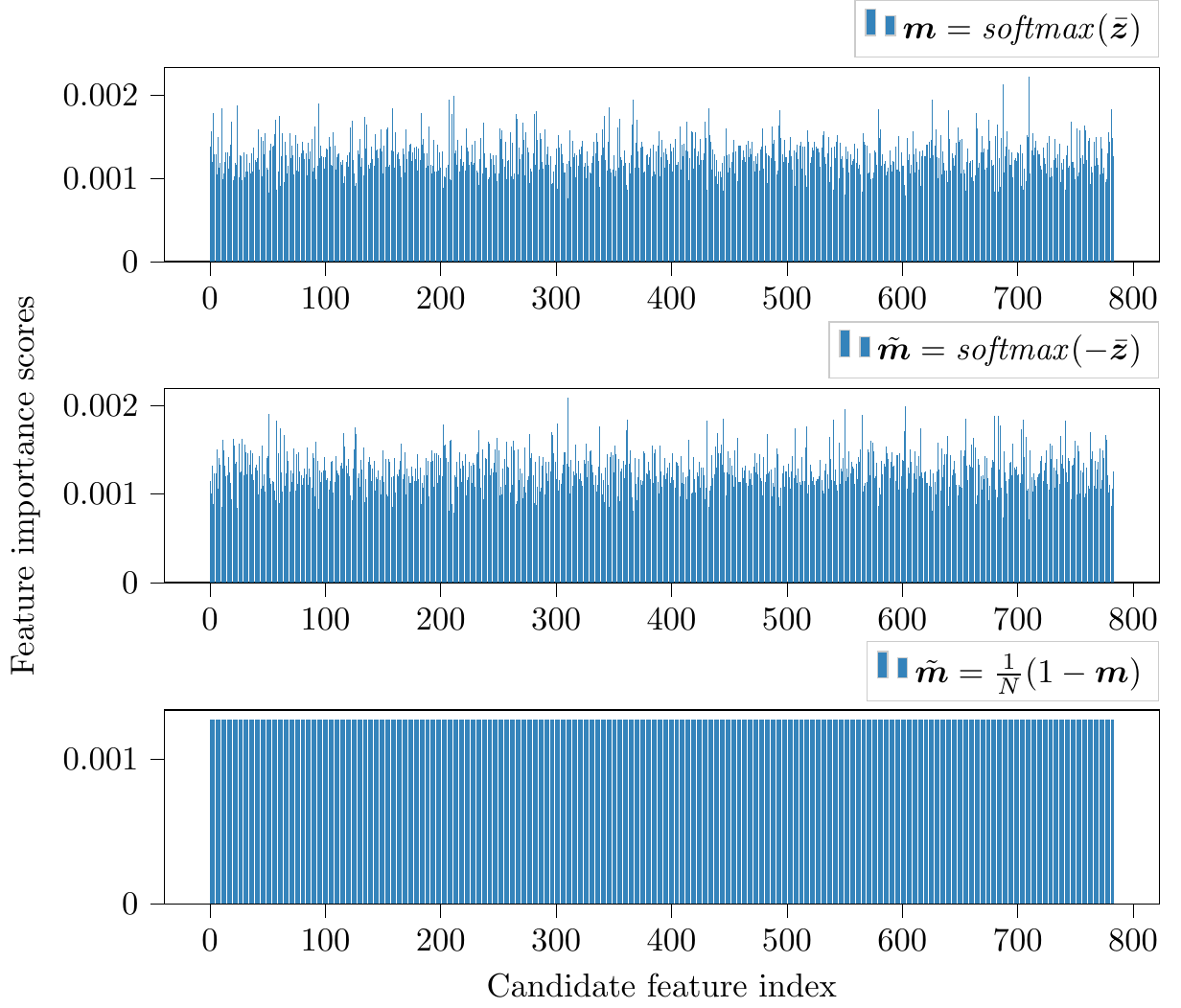}
		\caption{Initialization.}
		\label{fig:mask_initial}
	\end{subfigure}%
	\begin{subfigure}{.5\textwidth}
		\centering
		\includegraphics[width=1\linewidth]{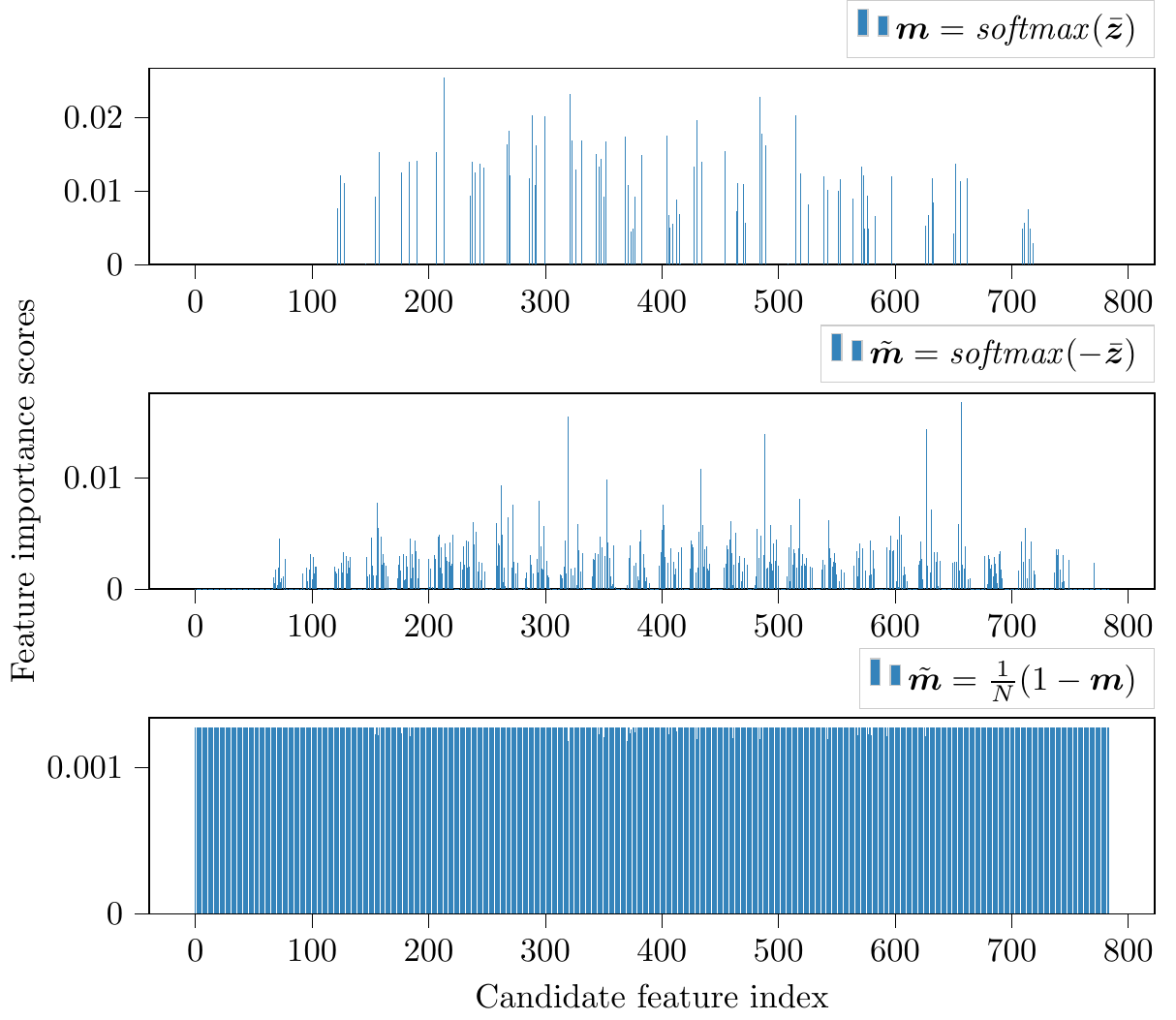}
		\caption{After training.}
		\label{fig:mask_trained}
	\end{subfigure}
	\caption{Feature importance for different designs of complementary feature masks. The proposed complementary feature mask $\tilde{\bm{m}}=\mathit{softmax}(-\bar{\bm{z}})$ can reliably assign notably different scores, while maintaining similar value ranges as the original feature mask $\bm{m}$.}
	\label{fig:test}
\end{figure}
As presented in Section~\ref{sec:method}, the CFM framework is generic and the design of the complementary feature mask can be specified by users and practitioners, enabling a lot of flexibility in practice. Nevertheless, an improper design of complementary feature masks might lead to difficulty in training. Therefore, this section visualizes the importance scores learned on the MNIST dataset to interpret how CFM defined in Equation~\ref{eq:coupled-mask} works. 

Figure~\ref{fig:test} shows the values of each element in $\bm{m}$ and $\tilde{\bm{m}}$ after initialization (Figure~\ref{fig:mask_initial}) and after training (Figure~\ref{fig:mask_trained}), respectively. In each sub-figure, the x-axis denotes different entries of $\bm{m}$ and $\tilde{\bm{m}}$ and the y-axis denotes the values of them. In particular, Equation~\ref{eq:fm} defines the feature mask $\bm{m}$ for the main path. Let the terms inside $\mathit{softmax}$ be denoted as $\bar{\bm{z}}$. We then have $\bm{m} = \mathit{softmax}(\bar{\bm{z}})$ and the complementary feature mask $\tilde{\bm{m}} =\mathit{softmax}(-\bar{\bm{z}})$. As can be seen from Figure~\ref{fig:mask_initial}, in our implementation, $\bm{m}$ (\emph{top}) and $\tilde{\bm{m}}$ (\emph{middle}) have similar value ranges. This is meaningful for initialization, meaning that all candidate variables can be almost equally considered at the beginning of training. Indeed, other design such as $\tilde{\bm{m}}=\frac{1}{N}(1-\bm{m})$ (\emph{bottom}) can also lead to similar value ranges. However, we found that such design also led to indistinguishable importance scores (around 0.001) after training as shown in the bottom of Figure~\ref{fig:mask_trained} and the scores were in a totally different value range than that of $\bm{m}$ (between around 0.01 and 0.02). On the contrary, our design can on one hand preserve similar value range for the complementary feature mask $\tilde{\bm{m}}$ shown in Figure~\ref{fig:mask_trained} (\emph{middle}) and the complementary scores of individual features are clearly discriminative.

In summary, our design has two benefits. Firstly, it allows similar value ranges for both $\bm{m}$ and $\tilde{\bm{m}}$ during initialization and training. Secondly, value differences between candidate features are obvious and forces the complementary path to focus on the features assigned with small scores by $\bm{m}$ during training.

\section{Discussion}
\label{sec:discussion}
As mentioned in the previous sections, the proposed CFM framework is generic. This means that the idea of using a complementary path to regularize the main learning task can be applied to other existing DL-based approaches. To briefly demonstrate this, we provide the pseudo-codes of applying CFM to a popular feature selection approach DFS~\cite{li2016deep}, where DFS does not contain a function $f_{\bm{\Theta}_m}(\cdot)$ denoted in Figure~\ref{fig:dl-fs}. This means that the feature mask $\bm{m}$ can be considered as a trainable vector which is randomly initialized. Accordingly, applying CFM to DFS is summarized in Algorithm~\ref{algo:dfs}. From the pseudo-codes, we can see that the CFM idea can be applied to existing DL-based FS method without much effort by simply adding a complementary path and the complementary loss $\mathcal{L}_c$. Meanwhile, the original loss functions remain without change. This is specially valuable for the use cases where some FS approaches have been used and the users do not want to significantly change the overall architecture and learning procedure.
\begin{algorithm}
	\caption{Applying CFM to DFS}
	\label{algo:dfs}
	\begin{algorithmic}[1]
		\REQUIRE Training dataset $\{(\bm{x}_1, y_1), (\bm{x}_2, y_2), \dots,(\bm{x}_N, y_N)\}$, the original loss function for DFS $\mathcal{L}_\text{DFS}$
		\STATE {Randomly initialize the DFS with the network parameters $\bm{\Theta}$ and $\bm{m}$}
		\FOR {$i = 1 $ to maximal training iterations}
		\STATE{Calculate the original loss as $\mathcal{L}_\text{DFS}(g_m(g_{fea}(\bm{x}\odot\bm{m})), y)$}
		\STATE{Calculate complementary feature mask as $\tilde{\bm{m}} = \mathit{softmax}(-\bm{m})$}
		\STATE{Calculate the complementary loss as $\mathcal{L}_c(g_c(g_{fea}(\bm{x}\odot\bm{m})))$}
		\STATE{Update $\bm{\Theta}$ and $\bm{m}$ based on gradient $\nabla_{\bm{\Theta}, \bm{m}}\mathcal{L}_\text{DFS} + \gamma\mathcal{L}_c$}
		\ENDFOR
	\end{algorithmic}
\end{algorithm}

Secondly, the choice or design for the loss function in the complementary path can depend on a given use case. In this work, we use categorical cross entropy w.r.t. a random class label as the loss function to force the complementary feature mask to have uncertain predictive capability and lead to better selection results. Nonetheless, we cannot exclude other implementations for loss functions. Some other intuitive ways can include maximizing the distance between the main and the complementary path, which is inspired from contrastive learning~\cite{chen2020simple}.

Finally, the complementary path is trained targeting an uncertain prediction. It may raise an question whether redundant relevant features can be selected. First, in terms of the prediction performance in downstream tasks, redundant features are not harmful and are acceptable, if it is allowed to select a feature subset with a moderately large size. In addition, as shown in Figure~\ref{fig:mnist_CFM_example} and Figure~\ref{fig:fmnist_CFM_example}, we found that the proposed framework with CFM regularization actually did not select redundant features (neighboring pixels), meaning that the neighboring pixels were assigned with notably different importance scores. This observation indicates that our method can automatically down-weight some redundant candidate features during training.
\section{Conclusion}
\label{sec:conclusion}
In this paper, we proposed a generic deep-learning-based feature selection framework based on a novel complementary feature mask. Based on a multi-task learning structure, CFM forced the candidate features with complementary importance to have as uncertain predictive ability as possible and thus enables a correct feature importance order in the main path. As a result, the selection quality of the main path was significantly improved over different selected feature subset sizes. Extensive experiments on six benchmarking datasets and comparison with a recent state-of-the-art method have shown the effectiveness and superiority of our method. In addition, we also demonstrated how to apply CFM to other existing approaches, showing that the CFM idea is generic. In total, this work is expected to inspire other fellow researchers to design new feature selection methods while considering the other side of the feature importance scores by using the complementary feature masks.

\section{Acknowledgment}
This research was supported by Advantest as part of the Graduate School ``Intelligent Methods for Test and Reliability'' (GS-IMTR) at the University of Stuttgart.

\bibliographystyle{elsarticle-num}
\bibliography{ref.bib}

\end{document}